\documentclass[final]{clv2025}

\jvol{vv}
\jnum{nn}
\jyear{2025}

\dochead{Short Paper} 

\pageonefooter{Action editor: \{action editor name\}. Submission received: DD Month YYYY; revised version received: DD Month YYYY; accepted for publication: DD Month YYYY.}

\usepackage{amsmath}
\usepackage{booktabs}
\usepackage{multirow}

\newcommand{\textdef}[1]{\textit{#1}}
\newcommand{\word}[1]{``#1''}
\newcommand{\citeposs}[1]{\citeauthor{#1}'s~(\citeyear{#1})}

\runningtitle{Running Article Title Here}
\runningauthor{Author's Surnames Here}

\begin{document}
\title{Relation Geometry \\in Semantic Space of Language Models}
\author{
Zhihan Cao\thanks{Corresponding authors}$^{,1}$,
Hiroaki Yamada$^{2}$,
Simone Teufel$^{3}$,
Tatsuya Hiraoka$^{4,5,6}$,
Kentaro Inui$^{4,7,5}$,
Hitomi Yanaka$^{1,5,7}$,
Takenobu Tokunaga$^{8}$
}

\affilblock{
    \affil{
    Graduate School of Information Science and Technology, The University of Tokyo\\\quad \email{zhihancao@g.ecc.u-tokyo.ac.jp, hyanaka@is.s.u-tokyo.ac.jp
    }
    }
    \affil{
    Faculty of Systems Design, Tokyo Metropolitan University\\\quad \email{ymd@tmu.ac.jp}
    }
    \affil{
    Department of Computer Science and Technology, University of Cambridge\\\quad \email{sht25@cam.ac.uk}
    }    
    \affil{
    Mohamed bin Zayed University of Artificial Intelligence\\\quad \email{tatsuya.hiraoka@mbzuai.ac.ae, Kentaro.Inui@mbzuai.ac.ae}
    }
    \affil{
    Center for Advanced Intelligence Project, RIKEN
    }
    \affil{
     Division of Information Science, Nara Institute of Science and Technology
    }
    \affil{
    Center for Language AI Research, Tohoku University
    }
    \affil{
    School of Computing, Institute of Science Tokyo\\\quad \email{take@c.titech.ac.jp}
    }
}

\maketitle

\begin{abstract}
When it comes to generating vector representations of words, current language models are achieving high-quality results.
However, what is not known is the extent to which knowledge about semantic relations is represented in the geometry of the semantic spaces created in this way. 
In order to answer this question, we study the \textbf{relation geometry} of such semantic spaces from three perspectives.
We first examine whether words standing in a particular relation to a target word~(called relata) occupy the same region in semantic space, and whether the regions corresponding to different relations are distinct from each other. 
We then verify to what extent semantic spaces reflect certain well-known properties of relations, such as symmetry, asymmetry, and transitivity.
Finally, we consider which information about the target words and relata is more important for relation geometry: their surface forms, or their contexts. 
We conduct experiments on six semantic relations using causal, masked, and diffusion language models. 
The results show that relata in asymmetric relations relatively clearly occupy a distinct region in semantic space. 
Asymmetric relations' properties are only moderately well encoded in the semantic space, yet better than those of symmetric ones. 
Furthermore, when considering the question which information source has the strongest impact on results amongst the models we evaluated, we find that lexical information tends to be more important for the causal language model, whereas contextual information is more important for the masked and diffusion language models.
Our results empirically show that relation geometry is not equally well-represented for all relations in semantic space, suggesting that there is a difference in how well semantic relations might be learned from distributional information alone.
This provides possible counter-evidence to the long-held distributional semantic view that the sense of words can be learned purely from their distribution.

\end{abstract}

\section{Introduction}
Transformer-based Language models~(LMs) achieve strong empirical performance in a wide range of tasks such as
word sense disambiguation~\citep{Mosolova_2025,Meconi_2025,Navigli_2026}, semantic change detection~\citep{Periti_2024,Umarova_2025, Desaetal_2026}, semantic similarity estimation~\citep{Zhou_2025, Liu_2025}, semantic parsing~\citep{Stengel_2023,Sherborne_2023}, and reading comprehension~\citep{Sauberli_2024}.
However, it remains unclear whether, and to what extent, their performance stems from a real understanding of language meaning~\citep{Cheng_2025}.
This concern has become a central topic in Computational Linguistics (CL).

The semantic theory upon which we rely when we talk about LMs is distributional semantics~\citep{Harris_1954}.
In distributional semantics, a linguistic unit's meaning is defined by its distribution alone.
According to \citeauthor{Harris_1954}, the distribution of a linguistic unit $w$ is the collection of $w$'s co-occurrent units in their particular positions.
In practice, the distribution of $w$ could be manifested by the sentences in which $w$ occurs.
Beyond the semantics of a single unit, we can also characterise the semantic relation between these units.
For this, the distributions of two units need to be considered, as follows.
\begin{quote}
\label{quote: harris}
    \textit{If the environments of A are always different in some regular way from the environments of B, we state some relation between A and B depending on this regular type of difference.
    [...]\\
     If A and B have some environments in common and some not (e.g. oculist and lawyer), we say that they have different meanings, the amount of meaning difference corresponding roughly to the amount of difference in their environments.
    \begin{flushright}
        \cite[][pp. 157]{Harris_1954}
    \end{flushright}
    }
\end{quote}
In short, if the distribution of $w$ differs from that of $v$ in a consistent way, then there should be a particular distributional relation between $w$ and $v$.
If $w$ and $v$ have near-identical distributions, they should be synonymy; but beyond synonymy, Harris did not discuss systematically how the distributional relations he defines map to semantic relations such as hypernymy and meronymy. 
Nevertheless, if Harris's theory is correct and such a mapping exists, then models based on distributional semantics should be able to capture semantic relations.

This motivates the need to examine such models, the results of which will provide an empirical foundation for better semantic theories.
Among all types of models, LMs are the one type currently widely used, with word embedding models~\citep[e.g.,][]{Mikolov_2013} in second place. 
Both have in common that they define a semantic space based on vectors that represent linguistic units.

If these models really capture semantic relations, then these semantic relations should be translatable to geometric relations in semantic space. 
We refer to this correspondence as \textdef{semantic relation geometry}.

A related notion is the linear representation hypothesis~\citep{Park_2024}, which proposes that semantic features in LMs are encoded as linear directions in their internal vector spaces. 
Under this view, features such as gender, sentiment, or part correspond to specific directions, and varying these properties amounts to moving along those directions.
What we develop in this article is an alternative view of semantics, where the space is defined by the location of the semantically related words of a target word in space.
We call the semantically related words \textdef{relata}.

\begin{figure}[hpt]
    \centering
    \includegraphics[width=0.5\linewidth]{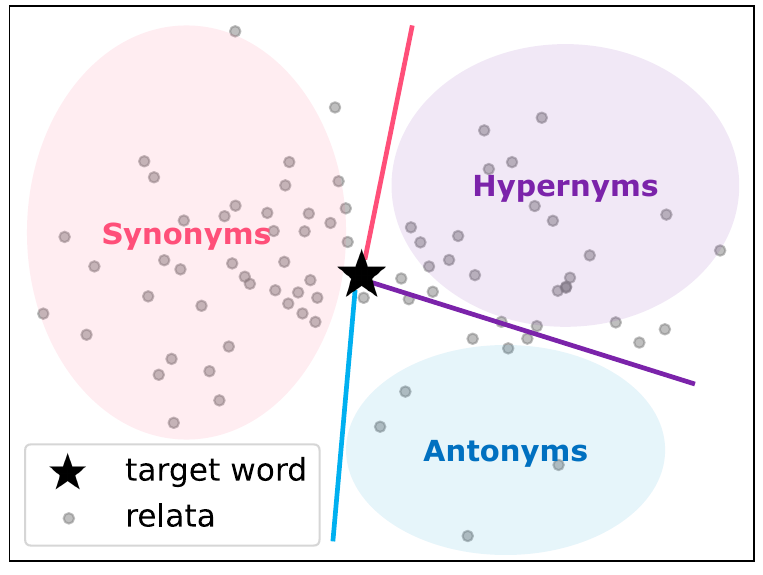}
    \caption{Illustration of relata regions.}
    \label{fig:region}
\end{figure}

Our working hypothesis is that, for a given target word, relata in the same relation occupy the same region, and relata in different semantic relations occupy linearly separable regions in the semantic space.
We call such regions \textdef{relata regions}, as illustrated in Figure~\ref{fig:region}.
If this is true, a simple prediction task should be able to reveal this correspondence: given a target word and a relatum in a particular relata region, we should be able to predict which relation applies. 

Semantic relation geometry concerns not only the existence of relata regions but also their properties. 
Each relation has a set of constitutive properties, which characterise the relation and differentiate it from other relations.
For instance, hypernymy and hyponymy are transitive.
If A is a hypernym of B and B is a hypernym of C, then A is a hypernym of C.
However, holonymy~(also called part-of relation) does not display transitivity.
Similarly, antonymy is symmetric: if A is an antonym of B, then B is also an antonym of A.
In contrast, hypernymy is known to be asymmetric.
Asymmetric relations usually have a {\em reverse relation}.
Hypernymy and hyponymy are reverse relations to each other, and so are holonymy and meronymy.

We expect the properties of a relation to have an observable effect on that relation's relata region, which allows us to verify the degree to which distributional semantics can capture semantic relations.
There is also a practical benefit of learning about properties of semantic relations: knowledge about such properties can directly support a wide range of reasoning tasks.
It can also provide explanations for the individual steps within the reasoning process. 

Our operationalisation of the concept of semantic space geometry is as follows.
We hypothesise a function that maps a pair of tokens' representations onto their relation.
We then train a relation probe $f$ to approximate such a function in a LM $M$'s semantic space.
We interpret the relation probe's performance as a measure of the extent to which our hypothesis on relation geometry holds. 

This article is motivated by two gaps in the current knowledge about LMs' representation of semantic relations. 
First, existing methods of studying semantic relations do not view semantic relations as a geometric operation, but we think that it is indispensable for understanding semantic relation geometry fully~\citep[e.g.,][]{Garcia_2021, Lin_2022}.
Second, of all relations, only hypernymy is well-studied~\citep[e.g.,][]{Ravichander_2020,Limisiewicz_2021}.

We develop a new interpretable probe-based method that can analyse semantic relation geometry, and we expand the evidence about semantic relations under study beyond hypernymy to hyponymy, holonymy, meronymy, synonymy, and antonymy. 
We also provide a full study of the extent to which relata regions reflect the properties~(symmetry, asymmetry, and transitivity) for all relations covered. 
In what follows, we compare different types of LMs (masked, diffusion, and causal LMs) as to their ability to distinguish and represent semantic relations along those lines.  

We are additionally concerned with another question. 
Since the sense of a token is determined by contextual information \citep{Harris_1954}, which contrasts with the lexical information contained in the token, we study the question of the relative importance of these two types of information for semantic relation geometry.
To do so, we develop a new methodology that quantitatively measures the importance of each regarding semantic relation geometry.

The article is structured as follows.
In Section~\ref{sec:related_work}, we summarise prior work about LMs and semantic relations.
In Section~\ref{sec:sem_method} and \ref{sec:sem_data}, we explain our probe-based methodology and the materials, comprising over 6.8 million triplets consisting of a target token, relation, and relatum.
Experiments and results follow in Section~\ref{sec:sem_results}.
We conclude the article in Section~\ref{sec:sem_summary} with limitations and future work.

\section{Related Work}
\label{sec:related_work}
Several empirical studies have concerned themselves with semantic spaces and semantic relations in connection with LMs.

\subsection{Probing For Hypernymy}
\citet{Ravichander_2020} presented one of the earliest such studies, establishing the basic methodology for the study of semantic relations. 
Their method is based on the probing, the use of classifiers to study linguistic regularities in semantic space\footnote{
See \citeposs{Belinkov_2022} survey on the history of probing in NLP}.
Given the LM representations of two tokens, the probe's task is to predict whether they stand in hypernymy relation or not.
The performance of the probe is then interpreted as how clearly hypernymy is encoded in semantic spaces.
The probe they used is a neural probe with a single hidden layer.
They also studied the effect of lexical leakage during probe training. 
In one setting, both the target token and its hypernym are unseen in training the probe; in the other, the hypernym is unseen but the target token is seen. 
WordNet provided the gold standard, and 576 pairs were tested in total. 
\label{sec:inflation}
The results showed that the probe performed close to the chance level (accuracy below 0.60) in the fully unseen condition, while achieving accuracy above 0.90 when the hypernym is unseen. 
\citeauthor{Ravichander_2020} concluded from this experiment that there is a danger of probe performance inflation, and that the semantic space created by BERT does not consistently reflect hypernymy.
This immediately raises the question whether other semantic relations might establish a more consistent semantic geometry.
Another limitation of this work is that it used only sense-unambiguous words (i.e., those with a single sense), but most words in the wild are polysemous. 

\citet{Limisiewicz_2021} also used probing to evaluate BERT's hypernymy knowledge against WordNet, but their task examines whether BERT representations encode structural information in the hypernymy hierarchy.
Specifically, they derive two hypernymy-related quantities: the depth of each word and the path distance in the WordNet hierarchy.
Their probe is linear and consists of two matrices: an orthogonal rotation matrix and a low-rank scaling matrix\footnote{
    They do not clarify whether they used monosemous or polysemous target words, which makes it harder to interpret their results.
}.
Their results showed that the two hypernymy-related quantities can be accurately predicted using only roughly twenty dimensions of BERT's semantic space. 
However, this result of a strong geometric relationship between hyponym and hypernym only holds for pairs of tokens co-occurring in the same sentence. 
Intersentential contexts were not considered by \citeauthor{Limisiewicz_2021}, but these are interesting because a) there are many such occurrences in the wild and b) in those cases, the syntactic structure and context might well be very different. 
We will address this gap. 

\subsection{Hypernymy Transitivity}
\citet{Aspillaga_2021} were the first to study a relation property explicitly, namely, transitivity of hypernymy.
To do so, they performed two tasks, hypernymy prediction and a newly invented task called hypernymy hierarchy reconstruction.
They created a hypernymy hierarchy across a set of candidate pairs, based on the likelihood of a pair standing in hypernymy, as predicted by a probe.
This way, the transitivity of hypernymy is directly modelled. 
Their probe is a Multi-layer Perceptron classifier.
They used 230,215 hypernymy pairs from Princeton WordNet Gloss Corpus \citep{wordnetglosscorpus}, a corpus of WordNet synset definitions where all open-class words are sense-disambiguated.
The gold standard for the hierarchy comes directly from WordNet. 

\citeposs{Aspillaga_2021} experiment has the advantage over~\citeposs{Ravichander_2020} that they cover polysemous as well as monosemous words. 
They studied three model families (masked LMs, causal LMs, and static embedding methods).
The results showed that, in general, masked LMs outperform the other types of LMs on both hypernymy prediction and hypernymy hierarchy reconstruction.

\citeposs{Lin_2022} study on hypernymy transitivity, which also uses sense-disambiguated words, used the sense's example sentences in WordNet.
In total, \citeauthor{Lin_2022} collected around 4,000 pairs, including direct and transitive hypernymy ones with non-hypernymy ones.
Their probe was a one-nearest-neighbour classifier, classifying whether a concatenated representation forms a hypernymy relation.
\citeauthor{Lin_2022} found that if two hypernymy pairs $(w,v)$ and $(v,u)$ are successfully classified as hypernymy, the transitive pair $(w,u)$ is classified correctly for about 80\% of the cases.

The results of the studies mentioned up to now, all of which used BERT to create the token representations, indicate that BERT acquires hypernymy transitivity to a certain extent.
The results of \citeauthor{Aspillaga_2021} and \citeauthor{Lin_2022} show that LMs may have learned transitivity, yet we still lack knowledge on either other relations or their properties.

\subsection{Other Relations}
Relations other than hypernym that were studied include synonymy. 
\citet{Garcia_2021} was the first to quantify how changes in contexts with respect to word senses influence synonymy.
\citet{Garcia_2021} replaced a target word $w$ in sentence $S_1$ with a synonym of the target word in the same sense and same context, $v$, creating sentence $S_2$, and compared this to $u$ in sentence $S_3$ that consists of the same context and a word $u$ with an entirely different sense, as follows:

\begin{quote}
    $S_1$: we’re going to the airport by \textbf{coach}.\\[1.1ex]
    $S_2$: we’re going to the airport by \textbf{bus}.\\[1.1ex]
    $S_3$: we’re going to the airport by \textbf{bicycle}.
\end{quote}
The working hypothesis is that if BERT encodes synonymy in its semantic space, the representation of $w$ in $S_1$ should be closer to the representation of $v$ in $S_2$ than the representation of $u$ in $S_3$. 
In other words, the inequality $\text{distance}(w,v)<\text{distance}(w,u)$ should hold consistently.
However, this inequality was satisfied in fewer than half of the cases.
When BERT was compared to the static embedding method, it was found that the static method satisfied the inequality in more than half of the cases.
\citeauthor{Garcia_2021} concluded that BERT does not reliably capture synonymy, possibly because it relies heavily on surrounding context and misrepresents words with different senses in similar sentences.

More recently, \citet{Pro_2025} studied meronymy in LMs' semantic space under the assumption of the linear representation hypothesis~\citep{Park_2024}, namely that the difference between the holonym and the meronym is encoded as direction in semantic space.
Using embedding and unembedding layers of ${\text{LLaMA-2-7b}}$~\citep{llama}, they created representations of around 1,000 meronym-holonym pairs.
They also created non-meronymy pairs. 
They then measured whether a meronymy direction vector exists.
If so, such a vector should maximise the inner product of itself and the vector differences of the meronymy pairs and minimise the inner product of itself and the vector differences of the non-meronymy pairs.
Such a meronymy vector was found only for specific categories, such as \word{vehicles} and \word{mammals}, suggesting meronymy is not linearly encoded uniformly across all meronym-holonym pairs.

\subsection{Limitations of Existing Studies}
The studies mentioned above have several limitations.
The first of these is empirical, as non-hypernymy semantic relations were left unexplored.
Some findings on hypernymy are insightful, but whether they can be transferred to other relations is unclear.
Similarly, there are no studies about properties of semantic relations,  with the exception of transitivity of hypernymy~\citep{Aspillaga_2021, Limisiewicz_2021}.
For example, we have no knowledge about the symmetry of synonymy and antonymy, or the asymmetry of hypernymy and holonymy.
As these properties provide an intrinsic characterisation of their respective relations, studying them will provide much-needed knowledge about whether LMs can learn \textit{how} two words are related, rather than just that they are related.

\label{sec:leakage}
The next two points concern methodology. 
In the previous studies by~\citeauthor{Aspillaga_2021} and \citeauthor{Lin_2022}, the probes learn hypernymy from both direct and indirect pairs at the same time, which means they can acquire transitivity relatively easily.
However, if we were to apply this to our situation, we would run into a potential information leakage problem.
We want to make sure that the probes are able to acquire a property in question without being exposed to explicit instances of that property.
To explain what we mean by such exposure, let us look at the example of hypernymy; the same principle applies to the other properties.
If two direct hypernymy pairs (A, B), (B, C), and one indirect pair (A, C) are in the dataset, then the probe's learning of hypernymy transitivity is trivial, as these three alone are sufficient to signal transitivity. 
A stricter and more realistic evaluation of transitivity would require that the entire dataset contains only direct pairs, so that no information about properties can possibly leak into the probe's learning. 

Another methodological limitation concerns the operation by which input representations into a probe are combined.
In the existing literature, this is either done by concatenation~\citep{Ravichander_2020, Aspillaga_2021, Lin_2022} or difference~\citep{Limisiewicz_2021, Garcia_2021, Pro_2025}.
However, concatenating two word representations yields a space that differs from the original semantic space of the LM studied, because it defines a space over token pairs rather than individual tokens.
These studies found no general hypernymy geometry, but it is possible that this may partly stem from the concrete realisations of the probe inputs used. 

There is also an empirical argument against using concatenation.
In experiments in computer vision and machine learning, \citet{Lei_2024} and \citet{Mahapatra_2025} found empirically that vector concatenation was a weak method for capturing interactions within paired representations of various entities. 
We are therefore suspicious that if applied to semantic relations, this operation might also underperform. 

We can borrow more sophisticated methods of modelling relations from knowledge graph embeddings~\citep[see the recent survey in][]{Cao_2024b}.
Nodes in a knowledge graph are represented as vectors, and relations between a target node and a relatum node are modelled as operations on these vectors.
One of the most straightforward approaches is to model relations via vector addition~\citep{Bordes_2013} or vector product~\citep{Sun_2019}.
Later approaches abstracted addition and multiplication into more general functions over the target and relatum nodes.
Such functions include linear mappings from a target node to a relatum node~\citep{Yu_2022} and bilinear functions that take both nodes as input and predict whether a relation of interest holds~\citep{Nickel_2016,Hayashi_2019-non}.

A bilinear function $\phi_r(w,v)$, defined as Equation~(\ref{eq:bilinear}), scores how likely two nodes in the knowledge graph ($w$ and $v$) are in relation $r$, based on suitable representations $e_w$ and $e_v$ of $w$ and $v$, respectively, and $W_r$ is the matrix that expresses to which degree relation $r$ holds between $w$ and $v$. 
Formally, it is 
\begin{equation}
    \phi_r(w,v) = e_w^TW_re_v \label{eq:bilinear}.
\end{equation}
\citeauthor{Bhattacharjee_2020} found that bilinear models capture cross-dimensional interactions far better than methodologies that model relations as vector addition or vector product.

\subsection{The Role of Contextual and Lexical Information}
An example is the work of \citeauthor{Garcia_2021}, who observed BERT's sensitivity to contextual information.
This sensitivity is not an isolated phenomenon and has also been observed outside the semantic relation domain~\citep{Zhuo_2024}.
We therefore need to look at how contexts contribute to the identification of semantic relations.

In addition to contextual information, lexical information has been shown to be theoretically and empirically important for some relations, such as antonymy~\citep{Justeson_1991, Cao_2025c}.
Furthermore, according to lexical semantic theory~\citep{Cruse_1986} and computational linguistics practice~\citep{Bevilacqua_2021}, sense disambiguation relies on both lexical and contextual information.
Lexical information provides the set of candidate senses, and contextual information resolves which one is appropriate.
Current LMs incorporate the lexical form of a token and the context in which it appears.
As a result, the representations they produce include both types of information and therefore should encode the sense precisely~\citep{Loureiro_2021}.
However, the roles of contextual and lexical information in semantic relations remain unclear.
Studying them could provide a more detailed understanding of LMs and inform theories of lexical semantics.

\subsection{Our Goals}
The current article aims to provide a holistic evaluation of how semantic relations are represented in LMs' semantic space.
To broaden the empirical scope, we study a diverse set of semantic relations: hypernymy, hyponymy, holonymy, meronymy, synonymy, and antonymy.
To achieve methodological interpretability, we use a probe that captures the interaction between representations and thus better characterises relational geometry.
We, unlike in previous work by \citet{Limisiewicz_2021}, study inter-sentential related pairs.
We replace existing simple probes based on the concatenation of vectors with a more sophisticated bilinear probe. 
When evaluating the performance of the probes, our criterion also addresses additional semantic relation properties. 
Finally, we propose a method to quantify the importance of contextual and lexical information.

\section{Method}
\label{sec:sem_method}
Recall that our working hypothesis is closely related to the probe performance $\mathrm{Perf}(f)$: high $\mathrm{Perf}(f)$ means that the semantic relation geometry is clear across semantic relations. 
We will now develop the evaluation methodology to do so, by first defining a bilinear probe for our purposes. 

\subsection{Probe}
The bilinear function in Equation~(\ref{eq:bilinear}) outputs a real number scoring how likely two objects stand in a relation of interest.
The probe $f$ we are developing needs to estimate the probability distribution over several relations instead.
We therefore apply the softmax as follows:
\begin{align}
    \label{eq:probe}
     f_r(w,v) &= \hat{\Pr}\left(r\mid w,v\right) = \frac{\exp\left(z_r\right)}{\sum_{r^\prime \in R}{\exp\left(z_{r^\prime}\right)}} \\
    z_r &= e_w^T W_re_v
\end{align}
where $e_w$ and $e_v$ are representations of $w$ and $v$, respectively, and $W_r$ is the relation matrix for $r$.
$R$ is a set of relations.
We obtain the probability distribution over relations by concatenating outputs of all $f_r$, as in Equation~(\ref{eq:whole_probe}).
\begin{align}
    \label{eq:whole_probe}
    f(w,v) &= [f_{r_1}(w,v),\ldots,f_{r_{|R|}}(w,v)]
\end{align}
The relation with the highest probability estimate is the prediction by the probe.

Bilinear models have a clear geometric interpretation.
Note that 
\begin{align}
    e_w^T W_r e_v = \left<W_r^T e_w, e_v\right>.
\end{align}
Consider $v$, which is a relatum that stands in relation $r$ with target word $w$.
A triplet $(w,r,v)$ then expresses that a target word $w$ and a relatum $v$ stand in relation $r$.
If we form the inner product of $e_v$ ($v$'s representation) with $W_r$, then this inner product should be larger than the inner product of $e_u$ with $W_r$, for any $u$ that is not in this relation with~$w$. 
Given $w$ and $r$, $f_r$ defines a hyperplane with normal vector ${W_r^T e_w}$.
$f_r$ expresses the degree to which relata and non-relata are linearly separable along the direction $W_r^T e_w$, that is, how many of $w$'s relata in $r$ lie on the same side of this hyperplane.
The full probe $f$, which considers all relations' hyperplanes at once, can be viewed as a set of linear separators based on~$e_w$.
It tests if it is possible to organise the relata of $w$ into relation-specific regions, relata regions, in semantic space.
According to our hypothesis, high probe performance on relation classification should indicate a situation where relata in different relations occupy different linear regions within the hyperplane set.

\subsection{Metrics}
\label{sec:sem_eval}
We define three metrics for evaluating probes.
{\em Relation Predictability} evaluates whether there is a clear relata region for a relation.
{\em Directionality} and {\em transitivity} evaluate the extent to which the relata region of a relation reflects a specific relation property.

In the upcoming experiments, we will perform $|I|$-fold cross-validation, where $I$ is the set of all folds.
The relation distribution over triplets is highly skewed, leading to large performance variations across folds and consequently unreliable estimates of the degree of relation geometry.
We perform $|I|$-fold cross-validation over $|J|$ repeated trials, where $J$ is the set of trials~(details in Section~\ref{sec:sem_data}).
We will explain below how this design interacts with the calculation of our metrics.

\subsubsection{Relation Predictability}
If token representation pairs in a given relation $r$ exhibit a clear relata region, then the probe $f$ should be able to predict the relations of the pairs in the test set.
We use the F1-score, the harmonic mean of precision and recall, where precision is the proportion of correctly predicted instances over all instances predicted as $r$, and recall is the proportion of correctly predicted instances over all instances whose relation is $r$\footnote{Full formulae in Appendix~\ref{app:fmeasure}.}.

For one of our folds $i$ from cross-validation, one of our trials $j$, and a given relation~$r$, we measure the relation predictability score $\mathcal{F}_r^{\{i,j\}}$.
Each trial will therefore achieve a trial-specific score $\mathcal{F}^{j}_r$ by averaging over folds as follows:
\begin{equation}
    \label{eq: trial_perf}
    \mathcal{F}_r^{j} = \frac{1}{|I|}\sum_{i \in I}\mathcal{F}_r^{\{i,j\}}
\end{equation}

We next derive relation-specific (Equation~(\ref{eq: relation_perf})) and summay relation predictabilities (Equation~(\ref{eq: general_perf})) as follows:
\begin{align}
    \label{eq: relation_perf}
    \mathcal{F}_r &= \frac{1}{|J|}\sum_{j \in J}\mathcal{F}_r^{j}\\
    \label{eq: general_perf}
    \mathcal{F} &= \frac{1}{|R|}\sum_{r \in R}\mathcal{F}_r.
\end{align}
Note that we applied macro-averaging over relations to arrive at $\mathcal{F}$.  
We decided against using micro-averaging because, again, the relation distribution in our upcoming experiments is extremely skewed, as we will see in Section~\ref{sec:sem_data}.

\subsubsection{Directionality} 
\label{sec:directionality}
We next define directionality as a binary feature expressing whether the order of $w$ and $v$ in a triplet $(w,r,v)$ influences its meaning. 
It operates differently for symmetric and asymmetric relations.
In symmetric relations, such as antonymy and synonymy, no influence of order should be observable because $(w,r,v)$ and $(v,r,w)$ have the same meaning. 
In asymmetric relations, such as hypernymy/hyponymy and holonymy/meronymy, there should be a strong influence of order.

We test the effect of order by exchanging the input order $(w, v)$. 
For a symmetric relation, we expect the prediction to remain the same, and for an asymmetric relation, we expect the prediction to indicate the reverse relation. 
For example, if $(w, v)$ stands in hypernymy relation, then we expect the probe to predict hyponymy for $(v, w)$.

Following this intuition, the directionality score for relation $r$, fold $i$, and trial~$j$,~$\mathcal{D}^{\{i,j\}}_r(f)$, is defined as:
\begin{align}
    \mathcal{D}^{\{i,j\}}_r (f) = \frac{1}{|T^{\{i,j\}}_r|} \sum_{(w_i, r, v_i) \in T^{\{i,j\}}_r}{
    \mathbb{I}\left[\arg\max f({w_i}, {v_i}) = h\left( \arg \max f({v_i}, {w_i})\right) = r
        \right],
    }
\end{align}
where $f$ is a probe, $T^{\{i,j\}}_r$ are those triplets in fold $i$ and trial $j$ which stand in relation~$r$.
$\mathbb{I}$ is the indicator function and $h$ maps a relation to its reverse relation, for example, $h(\text{synonymy}) = \text{synonymy}$ and $h(\text{hypernymy}) =\text{hyponymy}$.
The directionality score ranges from 0 to 1, with higher values indicating that directionality is better encoded in the relata region of the relation.
The same averaging procedure shown by Equation~(\ref{eq: trial_perf}), (\ref{eq: relation_perf}), and~(\ref{eq: general_perf}) is applied to directionality, to arrive at~$\mathcal{D}_r$, the relation-specific directionality score and ~$\mathcal{D}$, the summary directionality score per model.

\subsubsection{Transitivity} 
\label{sec:transitivity}
\textdef{Transitivity} is a property of hypernymy and hyponymy~\citep{Cruse_1986}.
For instance, the new hypernymy triplet (\word{robin}, HYP, \word{animal}) is derived from the two hypernymy triplets (\word{robin}, HYP, \word{bird}) and (\word{bird}, HYP, \word{animal}). 
Transitivity introduces the concept of distance $k$ between two senses, with direct triplets having a distance of one, and the distance of a derived indirect triplet being the sum of the distances of the triplets involved, with $K$ being the maximum distance observed\footnote{
The WordNet hypernymy hierarchy is a directed acyclic graph in which two senses may be connected by multiple paths~\citep{Richens_2008,Lohk_2014}.
This means that some triplets have ambiguous hypernymy distances. 
We will deal with this problem in Section~\ref{sec:indirect}.
}. 
Note that by the nature of transitivity, it can only be observed for indirect triplets, so we evaluate transitivity scores on indirect ones. 
$\mathcal{T}^{\{i,j\}}_{\{r,k\}}$ is calculated as the F1-score per fold $i$, trial $j$, relation $r$, and distance $k$, on the indirect triplets in that fold.

Based on this, we derive trial-specific transitivity scores $\mathcal{T}^{j}_{\{r,k\}}$ by averaging over folds, and distance-specific transitivity scores $\mathcal{T}_{\{r,k\}}$  by next averaging over trials, analogous to Equations~(\ref{eq: trial_perf}) and (\ref{eq: relation_perf}), with the exception of the additional parameter $k$. 
Averaging over distance achieves relation-specific transitivity score $\mathcal{T}_{r} = \frac{1}{k_{\text{max}}-1}\sum_{k=2}^{k_{\text{max}}}\mathcal{T}_{\{r,k\}}$, and summary transitivity score $\mathcal{T} = \frac{1}{|R|}\sum_{r\in R} \mathcal{T}_{r}$, where $k_{\text{max}}$ is the maximum distance.
Transitivity scores range from 0 to 1, with higher scores meaning that the relata region of relation $r$ reflects a higher degree of transitivity.

\subsection{Random Control}
A probe is itself a model, and its performance depends on two sources: the degree of semantic relation geometry in the semantic space under study, and the probe’s own ability to fit the label distribution in the training set. 
We follow \citeposs{Belinkov_2022} recommendation and train {\em random probes} (probes trained on random representations). 
As random probes learn only the label distribution, the difference between a real probe and its random version, which we call \textit{controlled} performance, allows us to cancel out the second effect.
We apply controlling after the final level of metrics we report, e.g., if we report $\mathcal{T}_r$, we will perform control at the relation level, but if we report $\mathcal{T}$, we will perform control at the macro-averaged level. 
We always report controlled performance, as it provides a clearer view of the portion of performance that reflects the semantic relation geometry.

The performance of random probes is obtained as follows.
For each token, we sample a random vector that has the same dimension as the token representations of a model to be controlled, and train it in the same way as the original representations.
The elements of the random vector are independently drawn from a uniform distribution on $[0,1]$.
Controlled performance can be negative when the probe fails to capture any relation geometry.
According to our hypothesis, in that case, relation-specific relata regions cannot be linearly separated in the semantic space, or no relational properties are present in the relata regions.

\subsection{Quantifying Contribution of Lexical and Contextual Information}
\label{sec:quantification}
We use ablation to examine how lexical and contextual information contribute to $f$ in the following way.
First, we remove lexical or/and contextual information from token representations; 
second, we train new probes on these manipulated representations; 
third, we measure the difference in performance $\mathrm{Perf}(f)$ of those new probes against the original probe.
These differences should reflect how important the lexical and contextual information are to $f$.
We will now describe the three steps. 

\subsubsection{Removal of Lexical Information}
Removal of lexical information is straightforward because the status of lexical information is binary: it is either present or absent.
This means that, for a sentence $s$ and a target token $w_i$, we only need to conceal $w_i$ while maintaining the rest.

Removal of lexical information is different for bidirectional and unidirectional LMs.
Bidirectional LMs predict a masked token using its contexts of both the left and right sides.
In this case, all we need to do is to replace $w_i$ with the mask token.
Because the mask hides the lexical information, its representation should now include only contextual information.
Unidirectional LMs predict a token using its preceding context.
The representation we need therefore must be based solely on the preceding context.
The representation of the preceding token $w_{i-1}$ meets this requirement and therefore is the representation we use. 

\subsubsection{Removal of Contextual Information}
Removal of contextual information is more complex because of the continuous nature of contextual information.
The attention mechanism dynamically creates the representation of a target token $w_i$ by adjusting the importance of each token in the context with respect to $w_i$.
This produces the attention score distribution of $w_i$, which contributes to a correctly contextualised representation.
Because of this entanglement between the attention mechanism and the contextual information, the contextual information cannot be removed without manipulating the attention mechanism itself. 

In principle, the attention score distribution of $w_i$ in a sentence $s=[w_1,\ldots,w_n]$ is given as:
\begin{align}
    \label{eq:attn}
    \alpha(w_i\mid s) = \mathrm{Softmax}\left(\frac{ K_s q_{w_i}}{\sqrt{d}}\right),
\end{align}
where $K_s \in \mathrm{R}^{n\times d}$ is the key matrix for $s$ and $q_{w_i} \in \mathrm{R}^{d}$ is the query vector for $w_i$, both computed by earlier modules\footnote{
Note that this is a simplification, as we omit both the layer and the attention head.
In practice, each layer of an LM has multiple attention heads, and each head produces its own attention score distribution for~$w_i$.
}.

We distort this distribution in order to obtain representations with contextual information removed. 
This can be achieved in two ways: make the attention score distribution uniform, or make the attention distribution one-hot for the target word.
Both manipulations switch off the attention mechanism and decontextualise the representations, but in different ways. 

In the one-hot case, which we call \textdef{degenerate decontextualisation}, the original context does not contribute at all to the representation, because all information about anything but the target word is simply removed. 

In the uniform case, which we call \textdef{egalitarian decontextualisation}, the representation of $w_i$ includes incorrect contextual information. 
Under this manipulation, relevant contextual tokens can no longer be distinguished from irrelevant ones, so all real information is lost. 

For degenerate decontextualisation, we simply replace each $\alpha(w_i\mid s)$ with an identity matrix whose diagonal elements are one and other elements are zero.

Egalitarian decontextualisation can be realised by introducing a temperature parameter $\tau$ into Equation~(\ref{eq:attn}):
\begin{align}
    \label{eq:temp_attn}
    \alpha(w_i\mid s, \tau) = \mathrm{Softmax}\left(\frac{ K_s q_{w_i}}{\tau \sqrt{d}}\right),
\end{align}
where $\tau \in (0,\infty).$
When $\tau$ approaches $\infty$, $\alpha(w_i\mid s,\tau)$ approaches a uniform distribution, and the result is the egalitarian-decontextualised representation.

\subsubsection{Lexical and Contextual Information Loss}
\label{pg:siv}
We denote the set of lexical information statuses by ${L = \{\mathrm{+}, \mathrm{-}\}}$, whose two elements stand for the presence and absence of lexical information, respectively.
The set of contextual information statuses is
${C = \{\mathrm{degenerate}, \mathrm{egalitarian}, \mathrm{correct} \}}$.

This means that in the upcoming experiments, there exist, for each metric, $2\times3=6$ controlled performance $\mathrm{Perf}_{\mathrm{ctrl}}$ for different combinations of lexical and contextual information status. 
A particular probe that learns representations with lexical information status $l \in L$, and contextual information status $c \in C$ is then expressed as $f^{l}_{c}$.

To allow for statements about how the two types of information (lexical vs. contextual) influence the evaluation metrics, summary statistics are needed. 

We define the loss from lexical information $\Delta \mathrm{LEX}$ as
\begin{align}
    \Delta \mathrm{LEX} =   
    \frac{1}{|C|}
    \sum_{c \in C}{
    \mathrm{Perf_{ctrl}}(f^{\mathrm{-}}_{c})
    - \mathrm{Perf_{ctrl}}(f^{\mathrm{+}}_{c})
    }.
\end{align}
$\Delta \mathrm{LEX}$ is the average performance loss from the absence of lexical information across contextual information statuses.
It quantifies the overall importance of lexical information for a given metric.

Similarly, the loss from correct contextual information $\Delta \mathrm{CTX}$ is defined as 
\begin{align}
    \Delta \mathrm{CTX} 
    =  \frac{1}{|L|}\frac{1}{|C|-1} 
    \sum_{l\in L}{\left(
        \sum_{c\in C \setminus \{\mathrm{correct}\}}{
        \mathrm{Perf_{ctrl}}(f^l_\mathrm{c}) 
        - \mathrm{Perf_{ctrl}}(f^l_\mathrm{correct}) 
        }
    \right)}
.
\end{align}
$\Delta \mathrm{CTX}$ is the average performance loss of correct contextualisations over incorrect cases.
It quantifies the overall importance of correct contextual information for a given metric. 

For each metric, relation, and LM, we compute $\Delta \mathrm{LEX}$ and $\Delta \mathrm{CTX}$. 
Smaller negative values of $\Delta \mathrm{LEX}$ and $\Delta \mathrm{CTX}$ indicate greater overall importance.
If, for one situation, both losses are negative and small, this would mean that lexical and contextual information are complementary. 
This is generally expected, as the two information sources are very different in nature. 
If, on the contrary, we would find that the two losses are both close to zero, this would indicate that neither piece of information is helpful. 
If one of the two is negative and the other is around zero, then information coming from the latter one can be recovered from the former one. 

\section{Data and Experiments}
\label{sec:sem_data}
We have introduced the methodology necessary for our experiments; we will now explain the materials used and the experimental setup.

\subsection{Target Words and Relata}
The source of target words and relata in our study is SemCor~\citep{semcor}.
SemCor provides manual sense disambiguation for all nouns, verbs, adjectives, and adverbs in 352 articles from the Brown Corpus~\citep{browncorpus}
WordNet senses were used for this, i.e., each token was assigned a synset tag.
WordNet establishes semantic relations between synsets, which are necessary for our experiments.

We restrict ourselves to nouns as target words and relata, because only nouns support the six relations of interest.
We exclude multiword expressions, named entities, and abbreviations because their interpretation relies on encyclopaedic knowledge or involves complex semantic compositional mechanisms, which introduce confounding factors and fall outside the scope of this study.
We also exclude sentences with fewer than five tokens because they may not provide enough contextual clues for LMs to determine word senses.
Additionally, the first words in a sentence are excluded as both target words and relata.
This makes sure that all LMs under study, whether they are unidirectional or not, have at least some context to work with.

After filtering, 68,228 tokens (10,953 types) remain, which correspond to 8,159 lemmas.
In total, the dataset contains 10,409 senses~(synsets).
For each token, we obtain its representation using LMs.
If a token is split into multiple subwords, we use the average of the subwords' representations as the token's representation.

\subsection{Triplets}
\begin{table}[th]
    \caption{Triplet statistics, by relation.}
    \centering\small
    \begin{tabular}{c|rrrr}
        \toprule
        Relation & Sense & Lemma & Type & Token \\
        \midrule
        HYP & 3,571 & 6,123 & 12,649 & 769,979 \\
        HPO & 3,573 & 6,127 & 12,750 & 753,224 \\
        HOL & 353   & 546   & 1,174  & 186,536 \\
        MER & 354   & 537   & 1,314  & 205,049 \\
        SYN & 1,637 & 2,496 & 4,594  & 194,674 \\ 
        ANT & 137   & 125   & 269    & 95,066 \\
        UNR & 500,000 & 493,500 & 496,419 & 500,000\\
        \midrule
        TOTAL & 509,625 & 509,454  & 529,169 & 2,704,528 \\
        \bottomrule
    \end{tabular}

    \label{tab:sp_pair_statistics}
\end{table}
We consider the six semantic relations: hypernymy~(HYP), hyponymy~(HPO), holonymy (HOL), meronymy (MER), synonymy (SYN), and antonymy~(ANT).
We will now describe the process of determining valid triplets, which takes the following steps:
\begin{enumerate}
\item Triplet collection: for each of the six relations, we extract triplets from SemCor, using WordNet information.
We include a control class, unrelatedness~(UNR), which contains pairs that do not stand in any semantic relation.
For UNR, we randomly sample 500,000 triplets from all unrelated triplets and make sure that none of them is related in any relation defined in WordNet\footnote{
Note that there is a minimal chance that the unknown triplets might stand in an actual relation, but only if the procedure happened to sample the target word and relatum of a real relation, and if that relation was missing from WordNet.
}.
\item Triplet cleaning: amongst all triplets, there are some that would cause problems of property leakage~(cf. Section~\ref{sec:leakage}). We remove those. 
\item Duplicated lemma filtering: additionally, if a synonymy triplet is trivial, namely contains the same lemma twice~(e.g. (\word{bird},  \word{birds}), we remove the triplet. 
\item Intra-sentential removal: we also exclude triplets whose token pairs co-occur in the same sentence, since co-occurrence may provide a shortcut for probes to learn relations, as may have happened in \citeposs{Limisiewicz_2021} experiment.
\end{enumerate}

As explained before on page~\pageref{sec:leakage}, we need to make sure that during the training of probes, no property under study is ever explicitly exhibited.
For transitivity, our evaluation only considers hyponymy and hypernymy pairs.
To perform the transitivity experiments fairly, we remove all sense pairs that do not stand in a direct hypernymy and hyponymy relation from the probe training set.
For directionality, when two sense pairs differ only in the order, we ensure that only one of the pairs appears in the relation mapping.
There are two cases. 
For symmetric relations, we need to treat the cases where two sense pairs are swapped.
For example, (\word{day}, ANT, \word{night}) and (\word{night}, ANT, \word{day}) cannot both be in the dataset.
For non-symmetric relations, we need to treat the cases where there are two sense pairs with swapped senses and reverse relations.  
For example, if (\word{robin}, HYP, \word{bird}) and (\word{bird}, HPO,  \word{robin}) cannot both be in the dataset.
We randomly discard one triplet in such cases.

Statistics at the sense, lemma, type, and token level are presented in Table~\ref{tab:sp_pair_statistics}.
In total, there are 3,571 sense pairs for hypernymy, 3,573 for hyponymy, 353 for holonymy, 354 for meronymy, 1,637 for synonymy, and 137 for antonymy.
At the token level, there are 769,979 triplets for hypernymy, 753,224 for hyponymy, 186,536 for holonymy, 205,049 for meronymy, 194,674 for synonymy, and 95,066 for antonymy.

\subsection{Training and Test Set Division}
\label{sec:lemma_split}
We use a lemma-based strategy to split the training and test sets.
Consider the set of all~$N$ triplets 
\begin{align}
    T = \left\{ (w_i,r_i,v_i) \mid i = 1,\ldots,N \right\}.
\end{align}
We denote the lemmatisation function of a token by $g$.
The sets of all $w_i$ and $v_i$ are called $W$ and $V$, respectively~
\footnote{$|W| = 10,867$, $|W| = 10,864$, $|W\cap V| = 10,778$.}
.
The set of lemmas that appear in all triplets of $T$ is 
\begin{align}
    M = \{g(u) \mid u \in W\cup V\}.
\end{align}
We make sure that no lexical leakage~(first observed by \citet{Ravichander_2020}, cf. Section~\ref{sec:inflation}) can occur. 
We split $M$ into two subsets, $M_{train}$ and $M_{test}$, in such a way that triplets from the two subsets do not share any lexical information.
In other words, the triplets of the training set contain only $w_i$ and $v_i$ with lemmas from $M_{train}$, and the triplets of the test set contain only $w_i$ and $v_i$ with lemmas from $M_{test}$.
To achieve this, we have to remove triplets where $w$ is in $M_{train}$ and $v$ is in $M_{test}$, or vice versa. 
For example, if (\word{body},~MER,~\word{leg}) is in the training set, then triplets that include either \word{body}, \word{bodies}, \word{leg}, or \word{legs} do not appear in the test set.
With this design, we ensure that probes are evaluated only on triplets whose tokens were not seen during training.

Probe training and evaluation are based on cross-validation, enabling the use of all triplets in probe evaluation.
To maintain lexical separation between training and test under cross-validation, we perform the training–test split for each fold at the lemma level as explained above.
The full lemma set $M$ is divided into $I$ subsets $\{M_1, ..., M_{I}\}$.
For fold $i$, the probe $f_i$ is trained on triplets $T_i = \{(w,r,v)\mid g(w),g(v) \in M\setminus M_i\}$, and then evaluated on $M_i$.
$\mathcal{F}$ and $\mathcal{D}$ scores are computed on each fold and averaged across folds.
Note that UNR is not a semantic relation; we do not expect it to display any relation property.
We therefore will not calculate $\mathcal{D}$ scores on UNR.

We do not stratify across target labels (here, relations) during cross-validation. 
As a result, the relation distribution for each fold can be skewed.
We repeat cross-validation over multiple trials in order to reduce the effect of skewness~(cf. Section~\ref{sec:sem_eval}).

\subsection{Indirect Hypernymy and Hyponymy Triplets}
\label{sec:indirect}
\begin{table}[thpb]
\centering\small
\caption{Statistics of indirect triplets, per distance.}
\begin{tabular}{c|rrrr}
\toprule      
Distance & Sense & Lemma & Type & Token \\
\midrule 
2  & 7,682 & 12,627 & 27,549 & 1,672,725 \\
3  & 7,921 & 12,325 & 28,057 & 1,688,907 \\
4  & 7,268 & 10,085 & 23,216 & 1,180,959 \\
5  & 5,871 & 7,199  & 16,822 & 910,786 \\
6  & 4,541 & 4,871  & 12,041 & 704,070 \\
7  & 3,432 & 3,415  & 8,755  & 417,708 \\
8  & 2,138 &  2,037 & 5,151  &  203,862 \\
9  & 1,232 &  1,145 &  2,866 &  101,881 \\
\midrule
TOTAL & 40,085 & 53,704 & 124,457 & 6,880,898 \\
\bottomrule
\end{tabular}
    \label{tab:sp_indir_pair_statistics}
\end{table}
The indirect hypernymy and hyponymy triplets which we removed from the training set constitute the sole test set for evaluating transitivity.
We create more hyponymy test data by reversing all indirect hypernymy triplets and merging them with indirect hyponymy triplets, and vice versa by adding reversed indirect hyponymy triplets to hypernymy triplets. 
This is done separately for indirect triplets of each distance.
As a result, the final sets of indirect hypernymy and indirect hyponymy triplets contain the same number of instances for each distance.

We discard triplets with distances greater than 10 because their accumulated counts are too low\footnote{
    This accounts for fewer than 1\% of all triplets.
} to ensure statistical reliability.
We also discard triplets whose distances are not unique\footnote{
    This causes a loss of 5\% of indirect triplets.
}, as they would endanger the interpretability of the results.
Table~\ref{tab:sp_indir_pair_statistics} reports the distribution of indirect triplets by distance.

\subsection{Models}
\begin{table}[ht]
\caption{Models used for generating token representations.}
\centering\small
\begin{tabular}{l l c c c}
\toprule
Model & Family & Layers & Dimensions & Parameters \\
\midrule
modernbert-large & MLM & 28 & 1024 & 395M \\
LLaDA-8B-Base    & DLM & 32 & 4096 & 8B \\
LLaMA-3.1-8B    & CLM & 32 & 4096 & 8B \\
\midrule
fastText~(baseline)        & Static & n.a. & 300 & n.a. \\
\bottomrule
\end{tabular}
\label{tab:sp_models}
\end{table}
We use three LMs to obtain token representations, each from a different family, as given in Table~\ref{tab:sp_models}. 
The first model is a masked LM~(MLM), modernbert-large, with 28 layers and 395 million parameters, producing 1024-dimensional token representations.
The second model is a diffusion LM (DLM) with 32 layers, LLaDA-8B-Base.
The third model is a causal LM~(CLM), LLaMA-3.1-8B.
Both LLaDA-8B-Base and LLaMA-3.1-8B have 32 layers and 8 billion parameters, producing 4096-dimensional token representations.
We use LLaMA-3.1-8B instead of larger variants in order to maintain comparability with LLaDA.
MLMs and DLMs are bidirectional LMs since they can take the preceding and following context of the target token into account; 
in contrast, CLMs are unidirectional and only take the preceding context into account.

The major component of an LM is a stack of Transformer layers.
Given a sentence $s = [w_1, \ldots, w_n]$, each layer of $M$ produces a vector representation for each token in $s$.
In this study, we use the output of the final layer as the representation of a token\footnote{
 We also examine the results using outputs from other layers; results are reported in the Appendix~\ref{app:layered_perf}.
}.

We use fastText~\citep{fasttext} as our baseline.
In fastText, the representation of a word is computed as the average of the representations of its constituent subwords.
Unlike representations produced by LMs, fastText representations are static, meaning that tokens with the same word type have the same representation independent of which context they appear in.
The dimensionality of fastText representations is 300.
We refer to modernbert-large, LLaDA-8B-Base, LLaMA-3.1-8B, and fasttext by ModernBERT, LLaDA, LLaMA, and Static, respectively.

\subsection{Experimental Details}
We train our probes using cross-entropy loss, optimised with Adam~\citep{adam}.
Training continues for five epochs with a batch size of 1,024.
Four-fold cross-validation over ten trials is applied.
In order to obtain degenerate and egalitarian decontextualised representations, $\tau$ is set to 100\footnote{
 We experimentally examined various values of $\tau$ and found that 100 is large enough to obtain a uniform attention score distribution.
}.

\subsection{Statistical Tests}
We perform two types of statistical tests.
The first is to determine, for a given metric and relation, which model performs best, using a paired permutation test with the controlled performances of different models in each trial as data points.
We perform the test only between a given model against the model with the numerically highest mean controlled performance among all remaining models.
We will show the best model per relation and metric in boldface in the result tables that follow. 

The second is to determine, for a given metric and model, which relation the model performs best on.
We first average performance across trials for each metric, model, and relation.
Then we apply the non-paired permutation test to the mean controlled performance on the relation being tested and the second highest mean controlled performance across all remaining relations, for each model and metric. 
We will underline the relation where a model performs best for a metric in the result tables that follow. 

For both types of tests, we use a one-sided alternative hypothesis and estimate p-values using Monte Carlo approximation with 1,000,000 permutations.
The significance level is set at $\alpha=0.01$. 
We apply the Bonferroni correction~\citep{Bland_1995} and set the significance level to $\frac{0.01}{120}$ per test.
This per-test significance level ensures that the probability of making at least one Type I error across all tests remains below $\alpha=0.01$.
The correction is necessary because we perform 120 statistical tests in total. 
This raises a multiple-comparison problem, whereby conducting multiple tests inflates the overall probability of Type I errors.

\section{Results}
\label{sec:sem_results}
In this section, we first present the performance of probes that learn representations containing lexical information and correct contextual information, and then report how lexical and contextual information influence probe performance.
We present the controlled performance throughout; the corresponding raw scores are provided in Appendix~\ref{app:raw_perf}.

\subsection{Relation Predictability}
\begin{table}[ht]
\centering\small
\caption{Controlled $\mathcal{F}_r$ scores across relations $r$ and models. Bold-faced values show the best model for the given relation, for each line. Underlined values show the best relation for that model, for each row.}
\begin{tabular}{c|cccc}
\toprule
Relation & Static & ModernBERT & LLaDA & LLaMA \\
\midrule
HYP & .12 & .23 & .19 & \textbf{.25} \\
HPO & .13 & .20 & .16 & \textbf{.22} \\
HOL & .14 & .17 & .17 & \textbf{.27} \\
MER & .12 & .15 & .14 & \textbf{.21} \\
SYN & .10 & \textbf{.12} & .08 & .10 \\
ANT & \textbf{.16} & .07 & .04 & .15 \\
UNR & \underline{.36} & \underline{.42} & \underline{.28} & \underline{\textbf{.46}} \\
\bottomrule
\end{tabular}
\label{tab:f1_controlled}
\end{table}
Table~\ref{tab:f1_controlled} presents the results of controlled $\mathcal{F}$ scores across relations and models.
We can see from the results that models do not perform well, remaining below 0.30 across the real semantic relations.
Models generally show higher scores on asymmetric relations~(hypernymy, hyponymy, holonymy, and meronymy) than symmetric relations~(synonymy and antonymy): the best $\mathcal{F}_r$ scores for the four asymmetric relations are in the 0.20 to 0.30 range, with the lowest one observed at 0.12. 
Best scores for the two remaining symmetric relations are 0.12 and 0.16, with the lowest one observed at 0.07. 

Models generally reach good performance on unrelatedness (e.g. LLaMA at 0.46, ModernBERT and the static baseline at 0.42 and 0.36).
This suggests that, given a token, the representations of tokens unrelated to it tend to occupy a distinct linear region in semantic space, which is more separable than those of the semantic relations. 

We also find that LLaMA clearly outperforms the other models on the four asymmetric relations.
LLaMA's advantage can also be seen in $\mathcal{F}$, the macro-average over the $\mathcal{F}_r$, which were measured at  0.16 for the static baseline, 0.15 for LLaDA, 0.20 for ModernBERT, and 0.24 for LLaMA in order of the value of $\mathcal{F}$\footnote{
We only report $\mathcal{F}$, $\mathcal{D}$, and $\mathcal{T}$ in texts.
}.
However, the advantage of LLaMA is not observed on symmetric relations; the static baseline performs best on antonymy~(0.16), and ModernBERT performs best on synonymy~(0.12).
 
In summary, all LMs can recognise asymmetric relations better than symmetric relations.
This suggests a reasonably linearly separable relata region for asymmetric relations, but no comparably separable region for symmetric relations.

\subsection{Directionality}
\begin{table}[ht]
\caption{
Controlled $\mathcal{D}_r$ scores across relations and models. 
Bold-faces and underlining are as before.
}
\centering\small
\begin{tabular}{c|rrrr}
\toprule
Relation & Static & ModernBERT & LLaDA & LLaMA \\
\midrule
HYP & .06 & \underline{.12} & .04 & \underline{\textbf{.18}} \\
HPO & \underline{.11} & \underline{.13} & \underline{.06} & \underline{\textbf{.17}} \\
HOL & .03 & .04 & .02 & \textbf{.07} \\
MER & .04 & .03 & .03 & \textbf{.07} \\
SYN & $-.$04 & $-.$02 & .01 & \textbf{.02} \\
ANT & \underline{\textbf{.12}} & .01 & .00 & .05 \\
\bottomrule
\end{tabular}

\label{tab:dir_ctrled}
\end{table}
Table~\ref{tab:dir_ctrled} presents $\mathcal{D}_r$ scores across relations and models.
We observe that no model reaches a score above 0.20 for any relation, with LLaMA reaching the maximum of all $\mathcal{D}_r$ scores for hypernymy, at 0.18.
The (macro-averaged) controlled $\mathcal{D}$ values, which do not exceed 0.10 for any relation, tell a similar story.
LLaMA again outperforms the other models, although this time it only reaches 0.09, with LLaDA reaching 0.03, and ModernBERT and the static baseline reaching 0.05. 

Again, asymmetric relations fare better than symmetric ones, with the maxima at 0.18 and 0.17 reached for hypernymy and hyponymy, respectively. 
However, in what looks like an outlier, antonymy reaches 0.12 with the static baseline while otherwise scoring low. 
Amongst the asymmetric relations, models generally perform better on hypernymy and hyponymy than on holonymy and meronymy.
$\mathcal{D}_r$ values for holonymy only range between 0.02 and 0.07, and for meronymy between 0.03 and 0.07.
Symmetric relations fare even worse, with synonymy even scoring negatively for the static baseline and ModernBERT (indicating performance below random chance), and where all scores remain below 0.05. 

In short, only hypernymy and hyponymy yield moderate performance for directionality.
Taking together the results of relation predictability, we can safely conclude that even though the relation predictability of asymmetric relations is of a moderate degree, their directionality is generally captured poorly by the LMs we evaluated.

\subsection{Transitivity}
\begin{figure}[ht]
    \centering
    \includegraphics[width=\linewidth]{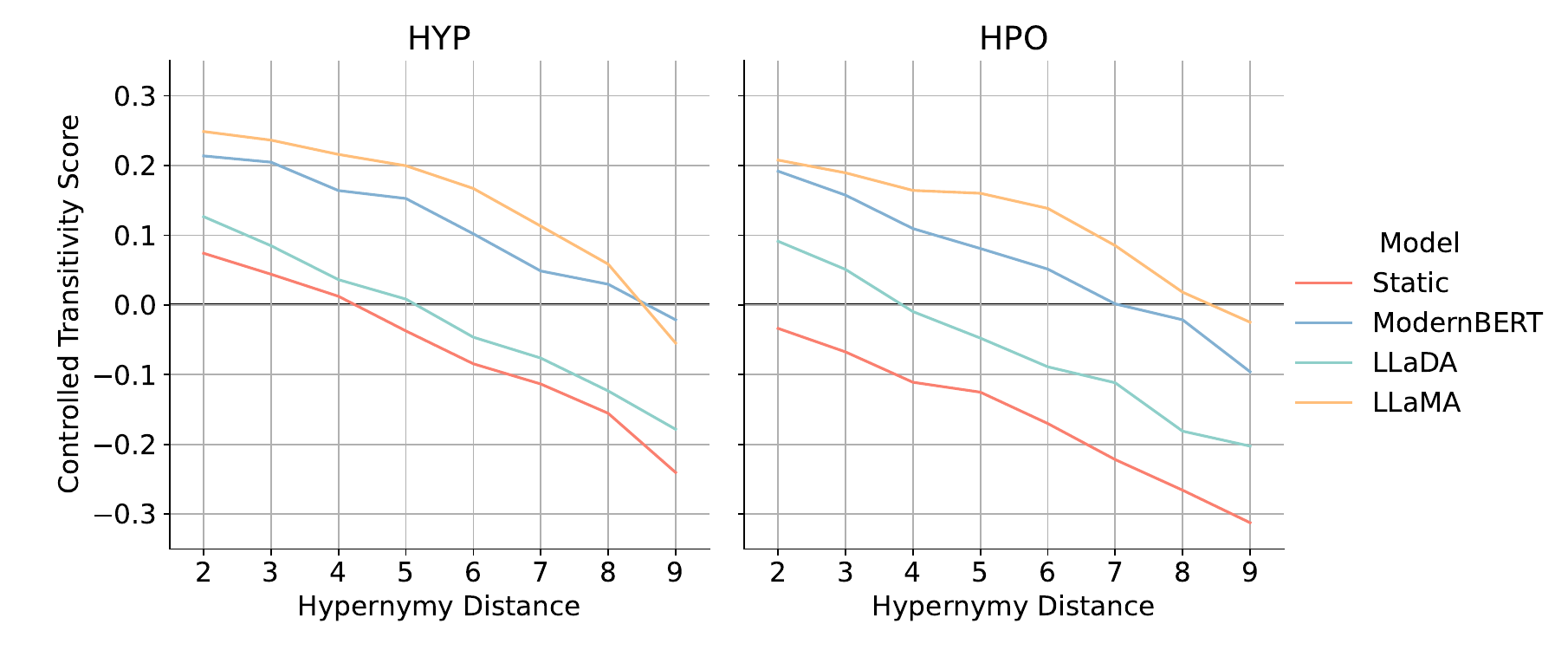}
    \caption{Controlled $\mathcal{T}_{\{r,k\}}$ scores per distance, for hypernymy and hyponymy.}
    \label{fig:sp_trans}
\end{figure}
Figure~\ref{fig:sp_trans} shows controlled transitivity scores $\mathcal{T}_{\{r,k\}}$ per distance~(2--9) for hypernymy and hyponymy.
Our first observation is that for all models, the controlled transitivity scores decline with increasing hypernymy distance.
The performance trends over distance are similar across models, with LLaMA achieving the highest scores for both relations at almost all distances (with the exception of hypernymy at the distance of nine). 
In particular, for hypernymy and hyponymy triplets with distances between two and six, LLaMA maintains scores above 0.10, numerically higher than the scores of other models.
However, for larger distances, LLaMA's performance declines and eventually falls below zero at a distance of nine.
The static baseline performs below chance at any distance for hyponymy, and after distance four for hypernymy.
LLaDA performs similarly badly, dropping below random chance before 5 for both relations.
  
\begin{table}[th]
\caption{Controlled $\mathcal{T}_r$ scores, averaged over hypernymy distances.}
\centering\small
\begin{tabular}{c|rrrr}
\toprule
Relation & Static & ModernBERT & LLaDA & LLaMA \\
\midrule
HYP & \underline{$-.$06 }& \underline{.11} & \underline{$-.$02} & \underline{\textbf{.15}} \\
HPO & $-.$16 & .06 & $-.$06 & \textbf{.12}\\
\bottomrule
\end{tabular}
\label{tab:sp_summary_trans}
\end{table}
Table~\ref{tab:sp_summary_trans} shows the $\mathcal{T}_r$ scores, which are averaged across distances.
LLaMA again outperforms other models over both relations.
LLaMA's advantage is observed in the summary $\mathcal{T}$ scores as well, where it achieves 0.13.
In contrast, ModernBERT achieves 0.09, whereas LLaDA and the static only obtain $-0.04$ and $-0.11$, respectively.

There is a general trend whereby all models perform better on hypernymy.
The $\mathcal{T}_r$ scores for the static baseline and LLaDA are negative, suggesting that for those models, the relation geometry does not reflect the transitivity of hypernymy and hyponymy.

In summary, performance on transitivity decreases with increasing hypernymy distance.
This suggests that indirect relations with long distances might have a different geometry from direct triplets in the semantic space.
As a result, for probes which learn only from direct triplets with no possibility of any leakage of transitivity, it becomes increasingly harder to perform well on such long-distance indirect triplets.

\subsection{Influence of Lexical and Contextual Information}
\label{subsec:lexctx}
Tables~\ref{tab:margidiff_f}, \ref{tab:margidiff_d}, and ~\ref{tab:margidiff_t} shows $\Delta \mathrm{LEX}$ and $\Delta \mathrm{CTX}$ for $\mathcal{F}_r$, $\mathcal{D}_r$, and $\mathcal{T}_r$ scores across relations and models, respectively. 
Lower deltas correspond to more influence of the factor concerned.
In these tables, boldface indicates for each model whether~$\Delta \mathrm{LEX}$ or $\Delta \mathrm{CTX}$ is lower.

\begin{table}[ht]
\caption{
$\Delta \mathrm{LEX}$ and $\Delta \mathrm{CTX}$ for relation predictability $\mathcal{F}_r$.}
\centering\small
\begin{tabular}{c||rr|r||rr|r||rr|r}
\toprule
\multirow{2}{*}{Relation}
& \multicolumn{3}{c||}{ModernBERT} 
& \multicolumn{3}{c||}{LLaDA} 
& \multicolumn{3}{c}{LLaMA} \\
\cmidrule(lr){2-4}
\cmidrule(lr){5-7}
\cmidrule(lr){8-10}
& $\Delta \mathrm{LEX}$ & $\Delta \mathrm{CTX}$  & $\mathcal{F}_r$
& $\Delta \mathrm{LEX}$ & $\Delta \mathrm{CTX}$   & $\mathcal{F}_r$
& $\Delta \mathrm{LEX}$ & $\Delta \mathrm{CTX}$  & $\mathcal{F}_r$
\\ 
\midrule 
HYP 
& $-$.11 & \textbf{$-$.15} & .23
& $-$.12 & \textbf{$-$.17} & .19
& \textbf{$-$.17} & $-$.06 & .25
\\
HPO 
& $-$.07 & \textbf{$-$.14} & .20
& $-$.04 & \textbf{$-$.10} & .16
& \textbf{$-$.17} & $-$.04 & .22
\\
HOL 
& $-$.03 & \textbf{$-$.13} & .17
& $-$.03 & \textbf{$-$.15} & .17
& \textbf{$-$.15} & $-$.10 & .27
\\
MER 
& $-$.03 & \textbf{$-$.12} & .15
& .00  & \textbf{$-$.13} & .14
& \textbf{$-$.08} & $-$.07 & .21
\\
SYN 
& $-$.05 & \textbf{$-$.06} & .12
& $-$.03 & \textbf{$-$.06} & .08
& \textbf{$-$.07} & $-$.03 & .10
\\
ANT 
& $-$.01 & \textbf{$-$.06} & .07
& $+$.01 & \textbf{$-$.06} & .04
& \textbf{$-$.07} & $-$.06 & .15
\\
UNR 
& $-$.22 & \textbf{$-$.23} & .42
& $-$.19 & \textbf{$-$.24} & .28
& \textbf{$-$.31} & $-$.16 & .46
\\
\bottomrule
\end{tabular}
\label{tab:margidiff_f}
\end{table}
We can see from Table~\ref{tab:margidiff_f} that ModernBERT and LLaDA prefer lexical information for relation predictability, whereas LLaMA prefers contextual information.
For LLaMA, the lowest losses are observed for unrelatedness, whose $\Delta \mathrm{LEX}$ is $-0.31$, followed by hyponymy and hypernymy at $0.17$.
For ModernBERT and LLaDA, $\Delta \mathrm{CTX}$ is lowest for hypernymy and followed by either hyponymy or holonymy.
In general, $\Delta \mathrm{CTX}$ is lower for asymmetric relations than for symmetric relations, being as high as $-0.06$ for antonymy and symmetry. 

\begin{table}[ht]
\caption{
$\Delta \mathrm{LEX}$ and $\Delta \mathrm{CTX}$ for directionality $\mathcal{D}_r$.}
\centering\small
\begin{tabular}{c||rr|r||rr|r||rr|r}
\toprule
\multirow{2}{*}{Relation}
& \multicolumn{3}{c||}{ModernBERT} 
& \multicolumn{3}{c||}{LLaDA} 
& \multicolumn{3}{c}{LLaMA} \\
\cmidrule(lr){2-4}
\cmidrule(lr){5-7}
\cmidrule(lr){8-10}
& $\Delta \mathrm{LEX}$ & $\Delta \mathrm{CTX}$  & $\mathcal{D}_r$
& $\Delta \mathrm{LEX}$ & $\Delta \mathrm{CTX}$  & $\mathcal{D}_r$ 
& $\Delta \mathrm{LEX}$ & $\Delta \mathrm{CTX}$  & $\mathcal{D}_r$
\\ 
\midrule
HYP & $-$.06 & \textbf{$-$.08} & .12 
& $-$.04 & \textbf{$-$.05} & .04
& \textbf{$-$.17} & $-$.02 & .18 \\
HPO & $-$.07 & \textbf{$-$.08} & .13 
& $-$.05 & \textbf{$-$.06} & .06
& \textbf{$-$.17} & $-$.01 & .17\\
HOL & .00 & \textbf{$-$.03} & .04 
& .00 & \textbf{$-$.03} & .02
& \textbf{$-$.04} & $-$.02 & .07 \\
MER & .00 & \textbf{$-$.02} & .03 
& .00 & \textbf{$-$.03} & .03
& \textbf{$-$.03} & $-$.03 & .07 \\
SYN & \textbf{$-$.03} & $+$.02 & $-.$02 
& $+$.05 & \textbf{$+$.03}  & .01
& \textbf{$-$.03} & $+$.02 & .02\\
ANT & \textbf{$-$.01} & \textbf{$-$.01} & .01 
& $+$.01 & \textbf{.00} & .00
& \textbf{$-$.04} & .00 & .05\\
\bottomrule
\end{tabular}
\label{tab:margidiff_d}
\end{table}
From Table~\ref{tab:margidiff_d}, we can see the same marked trend for $\mathcal{D}_r$ as $\mathcal{F}_r$, with ModernBERT and LLaDA showing a slight preference for lexical information and LLaMA generally preferring contextual information\footnote{
There is only one exception: this time, for ModernBERT, synonymy receives more benefit from lexical information.
For relations with small $\mathcal{D}_r$, we can observe that the $\Delta \mathrm{LEX}$ and $\Delta \mathrm{CTX}$ tend to be close to zero.
This is because, if the relata region reflects little directionality, the relative strength of contributing information is limited. 
}.
This effect is the most obvious for hypernymy and hyponymy.
For LLaMA, hypernymy and hyponymy profit strongly from lexical information at $\Delta \mathrm{LEX}$ of $-0.17$.
For ModernBERT and LLaDA, the influence of the two information sources is much more balanced for hypernymy and hyponymy~(with a 0.04 to 0.07 drop for lexical information removal versus a 0.05 to 0.08 drop for context information removal).

\begin{table}[ht]
\caption{
$\Delta \mathrm{LEX}$ and $\Delta \mathrm{CTX}$ for transitivity $\mathcal{T}_r$. }
\centering\small
\begin{tabular}{c||rr|r||rr|r||rr|r}
\toprule
\multirow{2}{*}{Relation}
& \multicolumn{3}{c||}{ModernBERT} 
& \multicolumn{3}{c||}{LLaDA} 
& \multicolumn{3}{c}{LLaMA} \\
\cmidrule(lr){2-4}
\cmidrule(lr){5-7}
\cmidrule(lr){8-10}
& $\Delta \mathrm{LEX}$ & $\Delta \mathrm{CTX}$  & $\mathcal{T}_r$
& $\Delta \mathrm{LEX}$ & $\Delta \mathrm{CTX}$  & $\mathcal{T}_r$
& $\Delta \mathrm{LEX}$ & $\Delta \mathrm{CTX}$  & $\mathcal{T}_r$
\\ 
\midrule
HYP 
& \textbf{$-$.10} & $-$.06  & .11
& \textbf{$-$.11} & $-$.02  & $-$.02
& \textbf{$-$.24} & $+$.07  & .15
\\ 
HPO 
& $-$.03 & \textbf{$-$.05} & .06
& \textbf{$+$.05} & $+$.09 & $-$.06
& \textbf{$-$.20} & $+$.08 & .12
\\
\bottomrule
\end{tabular}
\label{tab:margidiff_t}
\end{table}
From Table~\ref{tab:margidiff_t}, which covers transitivity, we observe that all models generally prefer lexical information, with the exception of ModernBERT for hyponymy.
When comparing Table~\ref{tab:margidiff_f} and Table~\ref{tab:margidiff_t}, we can make statements about how the different models handle direct vs indirect hypernymy and hyponymy, and which information they use to do this job. 
For LLaMA, for both indirect and direct relations, it is lexical information that is consistently more important.
LLaDA and ModernBERT, when it comes to direct hypernymy and hyponymy (Table~\ref{tab:margidiff_f}),  are different from LLaMA in that they rely more on contextual information.
However, when it comes to indirect hypernymy and hyponymy, they tend to be similar to LLaMA in that they rely more on lexical information.

\subsection{Discussion}
One persistent finding in our experiments is that models perform poorly on synonymy.
A possible reason is that we excluded trivial synonymy pairs, namely those sharing a lemma.
This makes the task challenging but allows us to study whether LMs capture synonymy at the semantic level.
Our finding provides counter-evidence to the conventional distributional hypothesis, which states that distributional similarity is proportional to semantic similarity, and that learning distributions yields knowledge of synonymy~(see the quote on page~\pageref{quote: harris}).

Combining the finding on synonymy with the observation that all LMs perform better on other semantic relations, particularly hypernymy and hyponymy, we can raise an important issue regarding distributions and semantic relations.
Our results partially support the distinction between semantic relatedness and semantic similarity, which has long been a concern in computational linguistics~\citep{Resnik_1995, Budanitsky_2006, Turney_2006, Hill_2015}.

Our experiments show that learning of distributions results in relatively clear semantic relation geometry of hypernymy, hyponymy, meronymy, and holonymy.
These are representative examples of semantic relatedness.
On the other hand, no such semantic relation geometry was found for synonymy, which is an obvious case of semantic similarity. 
This is consistent with the findings of~\citeauthor{Hill_2015}, who found that semantic similarity was harder to capture than semantic relatedness for distributional semantic models.
Note that in their experiments, co-occurrence-based and neural word embedding models are used.
Despite the architectural difference, both word embedding models and the LMs evaluated in this study learn purely from word distributions.
Therefore, this commonality suggests a potential limitation of distributional semantics in accounting for semantic similarity.

On antonymy, unlike synonymy, our static baseline outperforms the three LMs. 
One possible explanation is the fact that antonymy is not just a semantic relation but a lexico-semantic relation~\citet{Fellbaum_1995}.
In other words, what signals the antonymy is the lexical identity of the two antonyms, to a higher degree than it does for other semantic relations. 
The static baseline disregards context when representing words, but has access to particularly rich information about lexical identity, which might well result in the better performance of our baseline for antonymy.

Another finding confirms some aspects of~\citeposs{Sahlgren_2008} theoretical assumptions about the connection between relations and distributional semantics, while also revealing some aspects that should be further developed. 
\citeauthor{Sahlgren_2008} stated that models learning only the shared neighbours of two words (such as our baseline) should be better at paradigmatic relations~(e.g. semantic relations), whereas models learning the co-occurrence of two words should be better at syntagmatic relations~(e.g. collocations and idioms).
LMs learn both types of distributional information jointly, something that~\citeauthor{Sahlgren_2008} does not discuss.

The static baseline we used is based on skip-gram with negative sampling, which implicitly factorises a word–word pointwise mutual information matrix, as has been shown by~\citet{Levy_2014}.
This means our static baseline is a model accumulated from shared neighbours.
LMs, in contrast, learn not only shared neighbours but also co-occurrences.
The training objective of LMs requires them to predict a probability distribution over the vocabulary given a context.
As a result, LMs learn the likelihood of linguistic units co-occurring with a given context and learn to assign numerically close probabilities to alternative units in the same context.

The fact that our results show LMs’ relatively stronger performance on asymmetric relations over the static baseline suggests the presence of a synergistic effect of learning shared neighbours and learning co-occurrences, which seems to be particularly strong for asymmetric relations.
We can conclude that the effect of joint learning shared neighbours and co-occurrences, which has not been addressed in the previous theoretical discussions, requires more theoretical explanation and experimentation. 

When it comes to direct and indirect hypernymy relations, we also found some differences in the behaviour of the LMs studied that call for an explanation. 
The results in Table~\ref{tab:margidiff_f} and~\ref{tab:margidiff_t} support the claim that LLaMA relies more on lexical information than on contextual information for both direct hypernymy and hyponymy (corresponding to our tasks of relation predictability) and indirect hypernymy and hyponymy~(corresponding to our tasks of transitivity).
At the same time, ModernBERT and LLaDA prefer contextual information over lexical information for direct hypernymy and hyponymy, but prefer lexical information over contextual information for indirect hypernymy and hyponymy.

The differences in the behaviour of ModernBERT, LLaDA, and LLaMA for direct relations may lie in their respective language modelling objectives.
ModernBERT is a MLM, whereas LLaDA is a DLM.
MLMs and DLMs use the bidirectional context when predicting masked tokens, whereas CLMs such as LLaMA use the preceding context alone to predict the next token.
MLMs and DLMs thus learn the dependencies between the word to be predicted and both preceding and subsequent context, whereas CLMs learn dependencies of the word on preceding context together with the distribution of plausible subsequent context.
In other words, CLMs encode more varied information about the word, which might explain their larger performance drop observed when lexical information is removed.

The commonality in the behaviours of all LMs studied for indirect hypernymy and hyponymy may lie in the distributional difference of those relations from direct ones.
The context of a target word and the context of its relatum can be expected to be very similar to each other if the two tokens stand in direct hypernymy or short-distance indirect hypernymy. 
In contrast, a target word and a relatum in long-distance indirect hypernymy may exhibit little semantic overlap, and their contexts are likely to be dissimilar. 
In particular, the hypernym's context may be general and abstract.
Therefore, the contextual difference alone might suffice to identify direct, but not indirect, hypernymy or hyponymy.
Consider the following sentences from SemCor.
The target tokens are shown in boldface.
\begin{enumerate}
    \item The parallel bars, horse ... and mats formerly in the school \textbf{gyms} were replaced by [...] football.
    \item Some areas may already have been improved and contain \textbf{buildings} [...].
    \item In most cases, we recognise certain words, persons, animals or \textbf{objects}. 
\end{enumerate}
Given the context, it is relatively easy to identify the hypernymy relation holding between \word{gym} and \word{building}, which is direct hypernymy. 
On one hand, the contexts include words such as \word{in} and \word{contain}, which signal that both refer to physical structures.
This is different for indirect hypernym relation, like that holding between \word{object} and \word{gym}. 
Here, the available contextual cues are few, making the decision to recognise that this is hypernymy difficult.

\section{Conclusion}
\label{sec:sem_summary}
We have studied the geometry of semantic relations in the semantic space of LMs.
We addressed three questions:
first, whether relata regions for semantic relations exist;
second, whether such regions reflect well-known properties of relations;
and third, how important lexical and contextual information are to these regions.
Our contributions are as follows.
\begin{enumerate}
    \item We presented an interpretable evaluation methodology that approximates the relata regions of semantic relations in semantic space using probing.
    These probes were evaluated from three perspectives: how separable relata regions are; and the extent to which the relata regions reflect directionality and transitivity.
    \item We defined the statuses of lexical and contextual information and proposed methods, which we operationalise by manipulating the information going into token representations produced by LMs. 
    Based on these methods, we proposed novel evaluation metrics that quantify the importance of lexical and contextual information for semantic relations.
    \item Using these methods, we conducted an in-depth evaluation under a carefully designed experimental setting. 
     The evaluation covers six semantic relations, five of which are relatively underexplored, with unrelatedness as the control condition, and three LM families, including a DLM~(LLaDA-8B-Base), a MLM~(modernbert-large), and a CLM~(LLaMA-3.1-8B).
\end{enumerate}

The results showed that semantic spaces do not consistently distinguish between relations, whereas unrelated triplets can be linearly separated from related triplets. 
Models overall performed better on asymmetric relations than on symmetric relations.
Among asymmetric relations, hypernymy and hyponymy showed best results across metrics. 

We also found that the relata regions learned for those two relations reflect directionality to a moderate extent.
When it comes to transitivity, models were performing reasonably well for short-distance transitivity, but were generally unable to capture long-distance transitivity.

When comparing the LMs we tested, LLaMA performed better than LLaDA and ModernBERT in most cases.
However, on antonymy, our simple word embedding baseline model outperformed all LMs.

We designed an ablation method to quantify the importance of lexical and contextual information in the relata regions of semantic relations.
This was motivated by the observation that semantic relations are defined over senses and the sense of a word in running text is determined based on lexical information and contextual information, yet it is unclear how lexical and contextual information contribute to these relations.
In the experiments on this factor, we found that lexical information is consistently more important for LLaMA, whereas contextual information is more important for LLaDA and ModernBERT, except in the case of transitivity.

Our results provide an empirical basis for further discussion of distributional semantics. 
Learning the distributions of words does not guarantee the acquisition of semantic relations, which we equate to the emergence of relation geometry in semantic space.
Instead, our results call for further theoretical and methodological development in light of the contrast between our findings and the high performance of LMs in other tasks that require semantic knowledge. 

\section{Limitations and Future Work}
We used linear probes because linearity is highly interpretable and is suitable for the purpose of our evaluation.
This choice means that we cannot explore the question of whether semantic relations are represented non-linearly in semantic space.
The question of non-linear semantic spaces has recently become the topic of discussion~\citep{Valeriani_2023}, and there are good reasons to believe that our research could profit from non-linear probes.
After all, our probes perform extremely poorly on synonymy.
Future work should therefore explore more complex relata regions in semantic space.

Another limitation concerns the general interpretation of probing results.
Probes can detect knowledge encoded in representations, but this does not necessarily imply that such knowledge is actually utilised in downstream tasks~\citep{Belinkov_2022}.
In some cases, LMs may perform well on tasks that require specific knowledge even without explicit access to that knowledge.
We therefore do not make any claims here about the performance of LMs on tasks such as the construction of lexical resources~\citep{Santoro_2025,Bergh_2025}; the relationship between the knowledge probed and performance on such tasks needs to be empirically established independently.

When choosing LMs for this article, we did not attempt to address the full range of modelling approaches used in practice.
Instead, we only used three LMs, which we did not match for parameters or training data. 
It is possible that the observed superiority of LLaMA in our experiments is due to the fact that it has more parameters than ModernBERT and the static baseline, and that it is trained with a broader context window and on a larger corpus than LLaDA.
As LLaMA's performance advantage is likely due to the combined effect of these factors, our results do not allow us to conclude absolutely that CLMs capture semantic relations more effectively than other models.

Our choice of LMs follows from the purpose of this work, which is to study the limits of learning distributions alone, rather than engineering the best possible solution for the task of making LMs understand semantic relations.
We therefore only measure how a representative set of commonly used LMs would perform on our newly developed tasks. 
Models that incorporate multimodal information or alignment with human preferences, such as today's large proprietary LMs, may overcome some of these limitations, but examining such models falls beyond the scope of this article.

A further limitation of our work is that the experiments are conducted in highly controlled laboratory settings.
For example, we only study relations between nouns, rather than at the subword or multiword expression level. 
There are also limitations in our data: SemCor's sentences are drawn from the Brown Corpus, which reflects English before the 1970s and may differ from contemporary usage.
It is therefore imperative for the future to study semantic relations in more contemporary and realistic settings. 

\begin{acknowledgments}
This work was supported by JSPS KAKENHI Grant Number JP25KJ1271. 
\end{acknowledgments}

\newpage
\appendix
\appendixsection{Calculation of F1-score}
\label{app:fmeasure}
$\mathcal{F}^{\{i,j\}}_r$ for relation $r$ on fold $i$ in trial $j$ is calculated as follows.
\begin{align}
    \mathcal{F}^{\{i,j\}}_r &= \frac{2 \times R^{\{i,j\}}_r \times P^{\{i,j\}}_r}{R^{\{i,j}\}_r + P^{\{i,j\}}_r}\\
    R^{\{i,j\}}_r &= \frac{TP^{\{i,j\}}_r}{TP^{\{i,j\}}_r + FN^{\{i,j\}}_r}\\
    P^{\{i,j\}}_r &= \frac{TP^{\{i,j\}}_r}{TP^{\{i,j\}}_r + FP^{\{i,j\}}_r}
\end{align}
with $TP^{\{i,j\}}_r$ standing for true positives and $FN^{\{i,j\}}_r$ standing for false negatives within fold $i$ of trial $j$, while evaluating relation $r$.

\appendixsection{Raw Probe Performance}
\label{app:raw_perf}

We present here the performance across metrics, models, and relations of the raw probe (non-chance-controlled). 
\begin{table}[ht]
\caption{Performance across metrics, models, and relations\label{tab:raw_perf}}
\centering\small
\begin{tabular}{llccccccc}
\toprule
Metric & Model & HYP & HPO & HOL & MER & SYN & ANT & UNR \\
\midrule
\multirow{3}{*}{$\mathcal{F}_r$} 
   & ModernBERT & .42 & .41 & .20 & .19 & .08 & .15 & .53 \\
   & LLaDA & .41 & .38 & .22 & .18 & .05 & .12 & .47 \\
   & LLaMA & .49 & .46 & .32 & .26 & .15 & .15 & .62 \\
\midrule
\multirow{3}{*}{$\mathcal{D}_r$}  
    & ModernBERT & .17 & .18 & .04 & .03 & .03 & .01 & -- \\
    & LLaDA & .11 & .13 & .03 & .03 & .02 & .00 & -- \\
    & LLaMA & .25 & .25 & .07 & .07 & .03 & .05 & -- \\
\midrule
\multirow{3}{*}{$\mathcal{T}_r$} 
    & ModernBERT & .46 & .44 & -- & -- & -- & -- & -- \\
    & LLaDA & .37 & .32 & -- & -- & -- & -- & -- \\
    & LLaMA & .55 & .54 & -- & -- & -- & -- & -- \\
\bottomrule
\end{tabular}
\end{table}

\begin{figure}
    \centering
    \includegraphics[width=\linewidth]{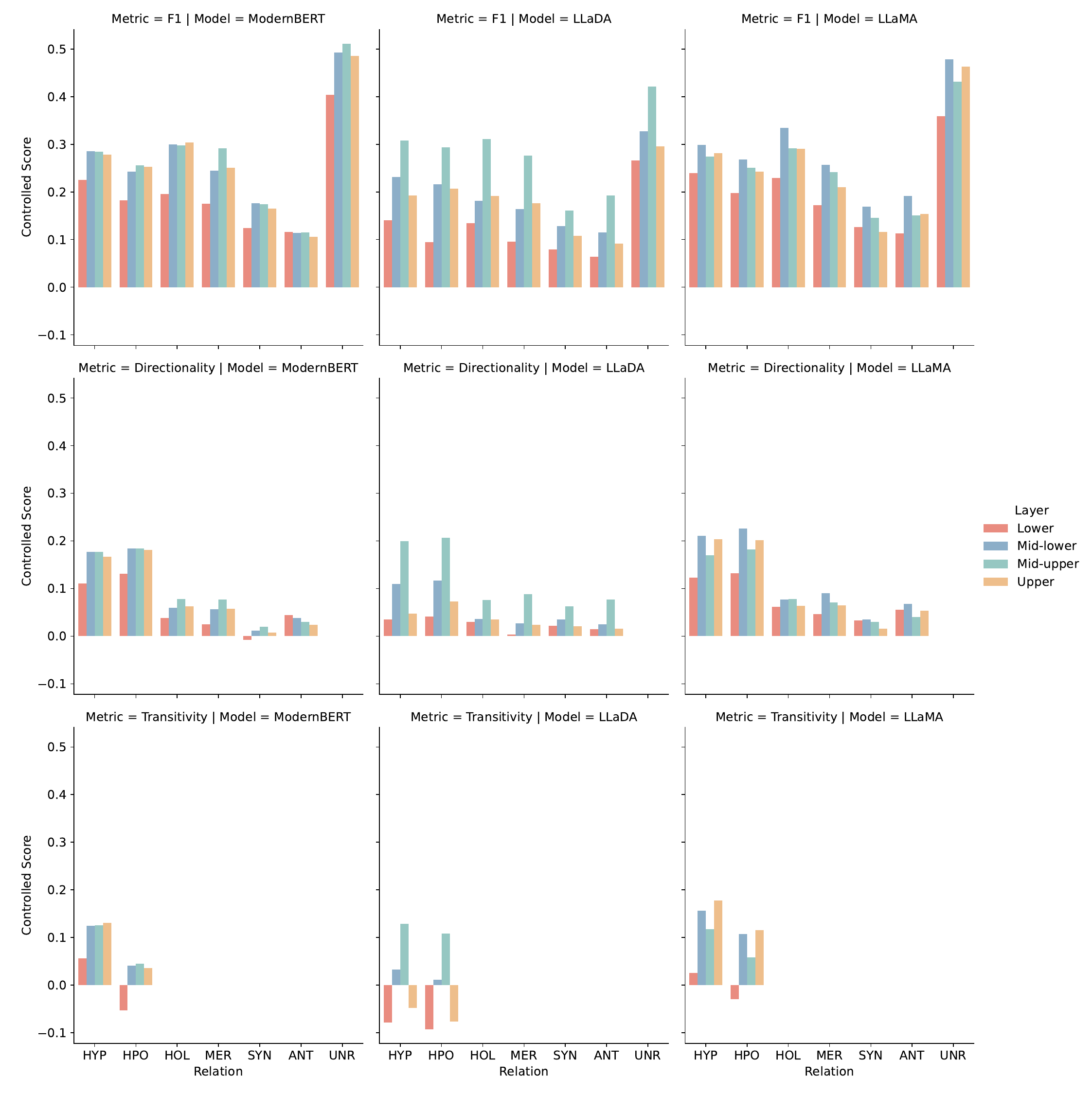}
    \caption{Controlled performance of different layers.}
    \label{fig:layered_perf}
\end{figure}

\appendixsection{Probe Performance on Representations by Different Layers}
\label{app:layered_perf}
Regardless of whether an LM is unidirectional or bidirectional, its layers have distinct functions and capture different types of information~\citep{Jawahar_2019,Rogers_2020,Aspillaga_2021,Merullo_2024}. 
Hence, it is necessary to examine how different layers contribute to the relation probe. 
However, the number of layers varies across LMs and cannot be directly compared. In practice, we divide the layers of each LM into four subsets: lower layers, mid-lower layers, mid-upper layers, and upper layers. 
Outputs within each subset are averaged to obtain token representations which reflect the information captured at different stages of the model. 
Figure~\ref{fig:layered_perf} shows the controlled performance of probes on representations from different layer subsets.

\newpage
\bibliographystyle{compling}
\bibliography{COLI_template}

\end{document}